\documentclass[runningheads]{llncs}

\usepackage[T1]{fontenc}
\usepackage{array}
\usepackage{graphicx}
\usepackage{tabularx}
\usepackage{url}
\usepackage {xcolor}

\usepackage{amsmath}
\usepackage{amsfonts}
\usepackage{amssymb}
\usepackage{caption}
\usepackage{subcaption}
\newcommand{\R}{\mathbb{R}}
\usepackage[skip=0pt plus0pt, indent=15pt]{parskip}

\begin{document}
\title{Segmentation tool for images of cracks}


\author{Andrii Kompanets\inst{1,3} \and Remco Duits\inst{2,3} \and Davide Leonetti\inst{1,3} \and Nicky van den Berg\inst{2} \and H.H. (Bert) Snijder\inst{1}}
\authorrunning{A. Kompanets et al.}
\institute{Eindhoven University of Technology, Department of the Built Environment, Eindhoven, the Netherlands \and Eindhoven University of Technology, Department of Mathematics and Computer Science, Eindhoven, the Netherlands 
\and Eindhoven Artificial Intelligence Systems Institute, Eindhoven University of Technology, the Netherlands\\
\email{a.kompanets@tue.nl \ d.leonetti@tue.nl}}

\maketitle

\begin{abstract}
Safety-critical infrastructures, such as bridges, are periodically inspected to check for existing damage, such as fatigue cracks and corrosion, and to guarantee the safe use of the infrastructure.
Visual inspection is the most frequent type of general inspection, despite the fact that its detection capability is rather limited, especially for fatigue cracks.
Machine learning algorithms can be used for augmenting the capability of classical visual inspection of bridge structures, however, the implementation of such an algorithm requires a massive annotated training dataset, which is time-consuming to produce. 
This paper proposes a semi-automatic crack segmentation tool that eases the manual segmentation of cracks on images needed to create a training dataset for machine learning algorithm. Also it can be used to measure the geometry of the crack.
This tool makes use of an image processing algorithm, which was initially developed for the analysis of vascular systems on retinal images. 
The algorithm relies on a multi-orientation wavelet transform, which is applied to the image to construct the so-called ‘orientation scores’, i.e. a modified version of the image.
Afterwards, the filtered orientation scores are used to formulate an optimal path problem that identifies the crack. 
The globally optimal path between manually selected crack endpoints is computed, using a state-of-the-art geometric tracking method.
The pixel-wise segmentation is done afterwards using the obtained crack path. 
The proposed method outperforms fully automatic methods and shows potential to be an adequate alternative to the manual data annotation. 

\end{abstract}

\section{Introduction}
There are more than a million bridges in Europe and the USA, according to \cite{brady2003cost,2Kitane2014} and every bridge structure is subjected to gradual deterioration which may lead to its collapse. 
In order to prevent tragic events and to detect critical structure deterioration in time, periodic inspections of bridges have to be conducted \cite{helmerich2007guideline}. 
Visual inspections are the most frequent type of bridge inspection, during which, trained personnel visually examines the surface of the structure \cite{rossow2012fhwa}. 
This procedure may be costly and time-consuming. 
Apart from the obvious expenses for inspector training and work, in some cases, there are implicit financial consequences to the regional economy related to the necessity to temporary restrict or fully shut down traffic across a bridge. 

In a recent paper from Campbell et al. \cite{6Campbell2020}, it has been highlighted that the variability of the outcome of visual inspections is substantial, and that the detection rate of fatigue cracks strongly depends on the inspection conditions and the proneness of the inspector to report potential damage. 
The authors concluded that, by using the current techniques and procedures, visual inspection should not be regarded as a reliable method for bridge inspection. 

The substantial cost and the low reliability of manual bridge inspections explain recent efforts directed at the designing of automatic systems for bridge inspections \cite{7Jo2018}. 
This is because automated visual inspection systems have the potential to augment the efficiency of existing inspection practices and to reduce its cost. 
In recent works, special attention was paid to automatic fatigue crack detection, since fatigue is one of the most frequent and dangerous types of bridge damage \cite{8Imam2012}.

Many computer vision algorithms have been developed for crack detection and measurement in different structures and materials, see for example overviews of this topic in \cite{9Mohan2018,10Ye2019}. 
These algorithms can be divided into two major groups, namely 
a) geometric image processing algorithms, and 
b) machine learning algorithms. 
Crack segmentation is a common approach extensively used in earlier works on automatic visual inspections, that besides crack detection, also allows measurement of the detected crack geometry. 
Image segmentation is the process of partitioning a digital image into multiple image segments, where each pixel is assigned to one of the segments. 
In the case of crack segmentation, two types of segments are considered: the segment of the image background type, and one or more segments of the crack type. 
Images of critical locations in bridge structures normally contain several elements that complicate the crack segmentation task. 
This may be surface corrosion, paint peeling, dirt on the surface, fasteners, and other bridge parts and joints. 
The performance of a fully automatic image processing algorithm dealing with such images is potentially affected by these aspects, e.g. it may result in a high false call rate by labeling a gap between adjoint elements of the considered structure as a crack, as mentioned in \cite{11Tong2021,12Yeum2015}. 
In contrast, computer vision machine learning algorithms not only can recognize geometrical patterns of cracks but can also distinguish them from other crack-like-looking parts of the image. 
However, the main drawback of machine learning algorithms is that they require a labeled dataset for training and testing. 
A large number of researchers developing machine learning algorithms for crack segmentation uses labeled datasets created by manual labeling. 
In this case, the annotation consists of a manual draw of the contour of the crack, most often made with a hand-held mouse \cite{Kim2020,Qiao2021}. 
Clearly, such a procedure may take a lot of time, especially in the case of a large number of high-resolution images. 

Taking this into consideration, a manual annotation procedure forms a bottleneck in the process of development of a state-of-the-art machine learning algorithm for crack segmentation. 
In this paper, we propose to make use of image processing techniques to develop a semi-automatic crack segmentation tool. 
The developed tool has the advantage of making the process of labeling cracks on images simpler and faster, which is needed to train a machine learning algorithm for crack segmentation. 
Moreover, the proposed tool may also be used to extract the geometrical parameters (length, width, curvature, etc.) of a crack needed  for either research or more accurate inspection.
Unlike the fully automatic image processing technique for crack segmentation, the developed method requires manual input, i.e. the crack end-points. 
By adding such a manual input it is possible to significantly increase the accuracy of the crack segmentation algorithm, potentially allowing it to be used as a data labelling tool for machine learning algorithms. Implementation of the developed algorithm is available at https://github.com/akomp22/crack-segmentation-tool

\section{Methods}
\label{sec:Methods}
Datasets to train a machine learning algorithm for crack segmentation are often manually labeled, since it is the most reliable method for pixel-wise image labeling. However, it is time consuming, because it requires a human expert to manually indicate the contours of the crack.

Fully automatic image processing algorithms do a crack segmentation much faster and without human involvement.
A wide range of different image processing techniques have been created aiming at the development of a fully automatic crack segmentation algorithm.
Commonly used techniques include thresholding \cite{16Talab2016,17Li2008,18Zhang2014,19Li2018,20Oliveira2014}, filtering \cite{22Vidal2016,23}, and image texture analysis \cite{24Huang2006,25,26Hu2010}.

\begin{figure}
\centering
\includegraphics[width = 0.9\textwidth]{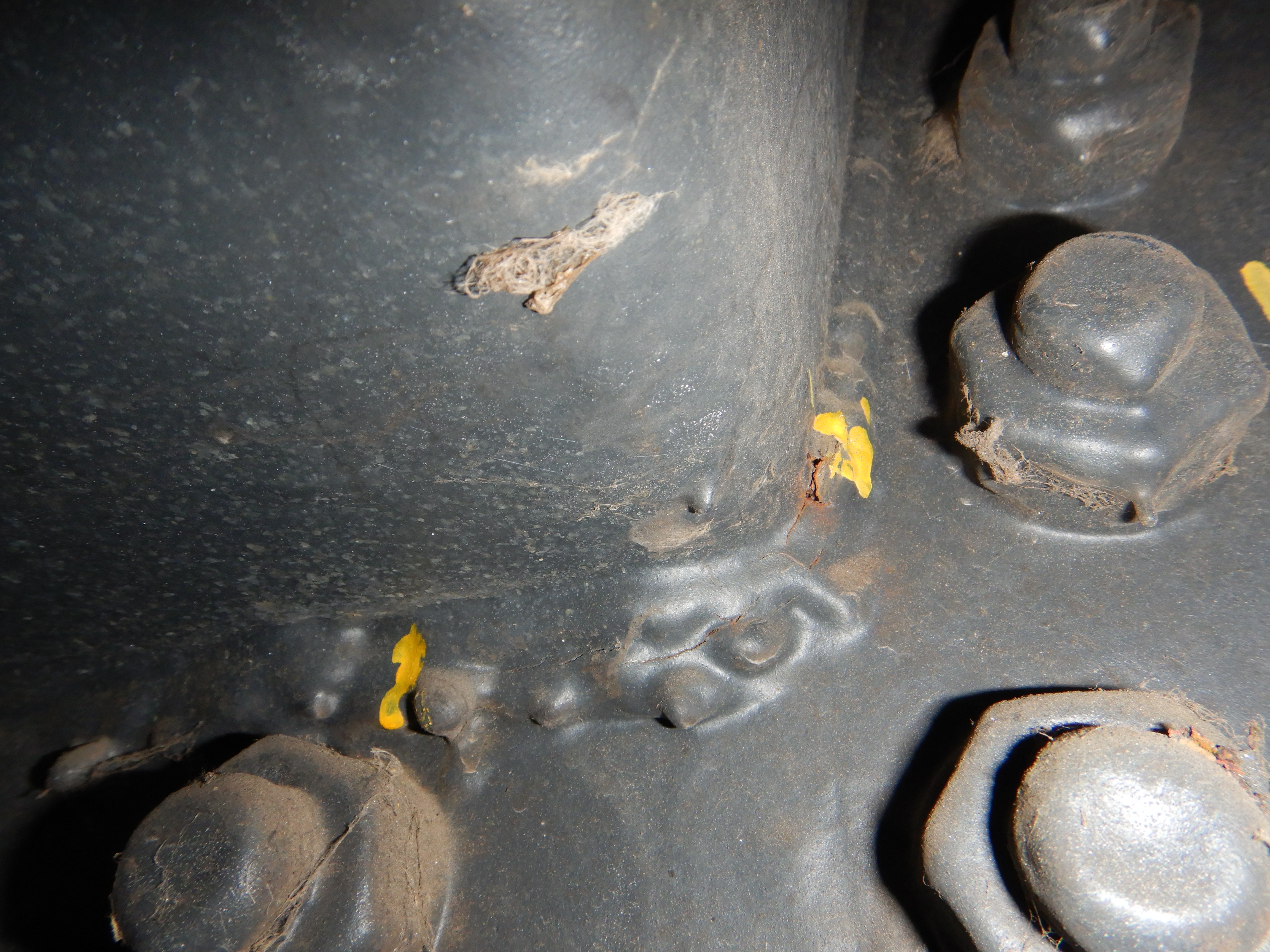}
\caption{ Example of a steel bridge structure image where common crack detection algorithms tend to have a high false call rate.}
\label{fig:Noisy steel bridge image}
\end{figure}

A major drawback of these fully automatic algorithms is that they are often designed to mark dark elongated structures of the image as a crack.
On complex images, for example the image of a steel bridge presented in the Fig. \ref{fig:Noisy steel bridge image}, these algorithms will suffer from the significant amount of false crack detections.
Hence, these algorithms can not be used as an image pixel-wise labeling tool.

The semi-automatic algorithm described in this section uses an image processing technique, where the false call rate is minimized thanks to additional information about the crack location provided by a human annotator. 

The first major step of the proposed method is crack path tracking, described in  Section \ref{sec:Crack path tracking} of this paper.
The algorithm aims at finding the path of a crack between two manually selected crack endpoints. 
The retrieved crack path allows to distinguish a particular elongated dark structure of the image (which is a crack) form other elements in the image that look like a crack (e.g. shading, paint, structure edge etc.).
Few earlier works considering crack tracking having selected points as input.
For instance, in \cite{dare2002operational}, a crack path was identified by a step-by-step movement from one of the crack points in the direction that has a locally minimal pixel intensity.
A disadvantage of this tracking method is that it is not able to find a globally optimal path.
It can easily be mislead by dark image regions near the crack and follow a locally optimal wrong path.

In \cite{kaul2011detecting} the crack path was found using only one internal crack point as input.
A fast marching algorithm was used to find a distance map relative to the selected internal point. 
The lowest gradient ascent path was chosen as a crack path. 

Crack minimal path selection approaches was also used in \cite{amhaz2014new,amhaz2016automatic}.
In these articles, Dijkstra's algorithm was used to find a crack path between multiple points inside crack contour. 
However, points inside crack contour were selected automatically, using the threshold method, so the problem of the high amount of false detections that is inherent to fully automatic image processing algorithms is also present in these methods. 

The proposed crack tracking method uses an anisotropic fast marching algorithm to find a crack path as a globally  optimal path (minimizing geodesic) in an orientation score (a multi-orientation wavelet transform) and is described in Section \ref{sec:Crack path tracking} of this paper. 

To obtain a crack segmentation, the width of the crack along the retrieved path needs to be measured.
One way to do this is to get an intensity profile perpendicular to the crack path, and determine crack edges based on this profile, as was done in \cite{dare2002operational}. However, the crack edges retrieved with this method have irregular shapes and may deviate significantly from the actual edges.
A width expansion approach was used in \cite{amhaz2014new,amhaz2016automatic,chen2021improved}, and is also used in the proposed algorithm and explained in Section \ref{sec:Crack width detection}.
An alternative width measurement method is also proposed (tracking algorithm for edges as described in Section \ref{sec:Crack width detection}) which works better on images of cracks in steel bridges.

\subsection{Crack path tracking}
\label{sec:Crack path tracking}
\begin{figure}
     \centering
     \begin{subfigure}[b]{0.49\textwidth}
         \centering
         \includegraphics[width=\textwidth]{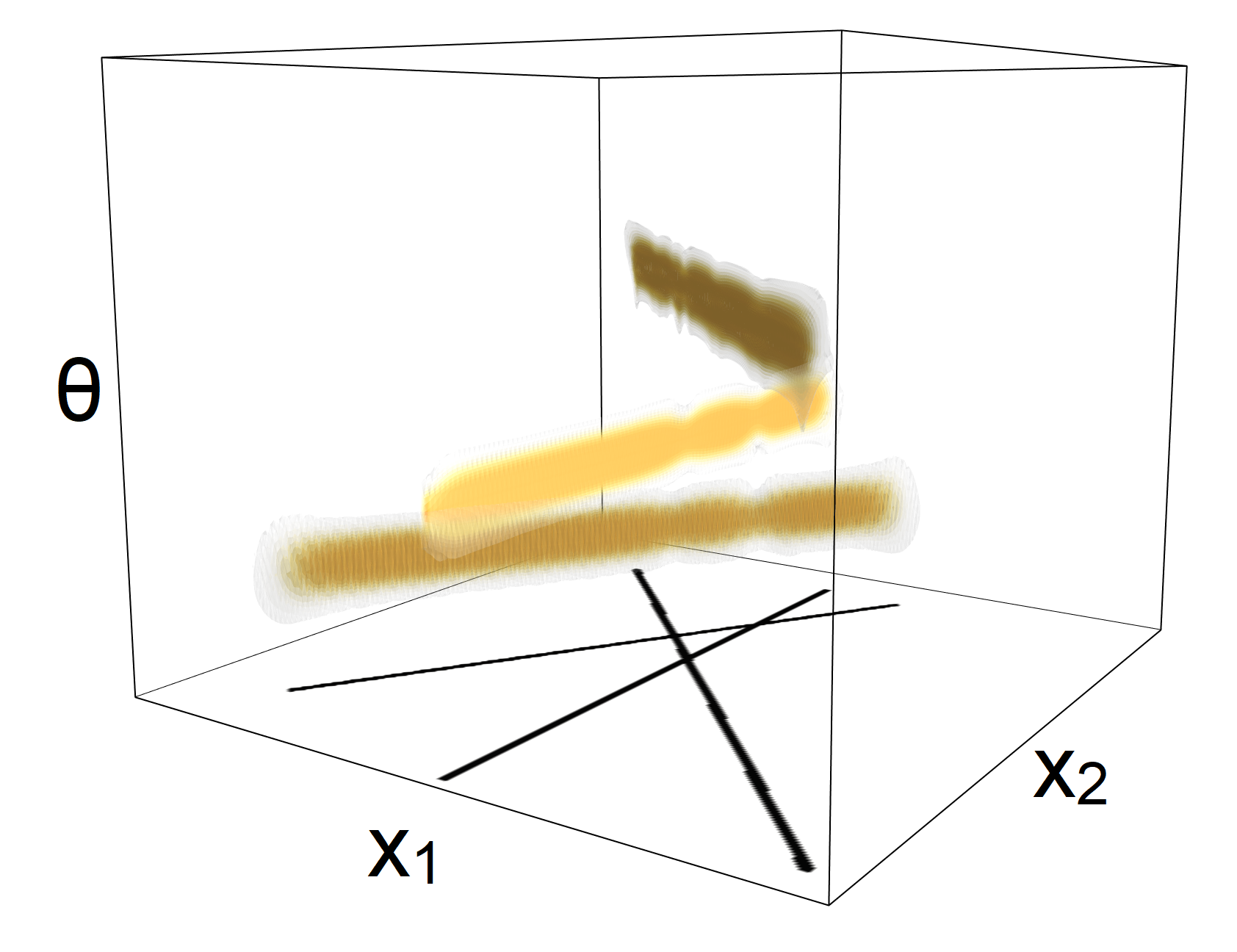}
         \caption{}
         \label{fig:os_lines}
     \end{subfigure}
     \hfill
     \begin{subfigure}[b]{0.49\textwidth}
         \centering
         \includegraphics[width=\textwidth]{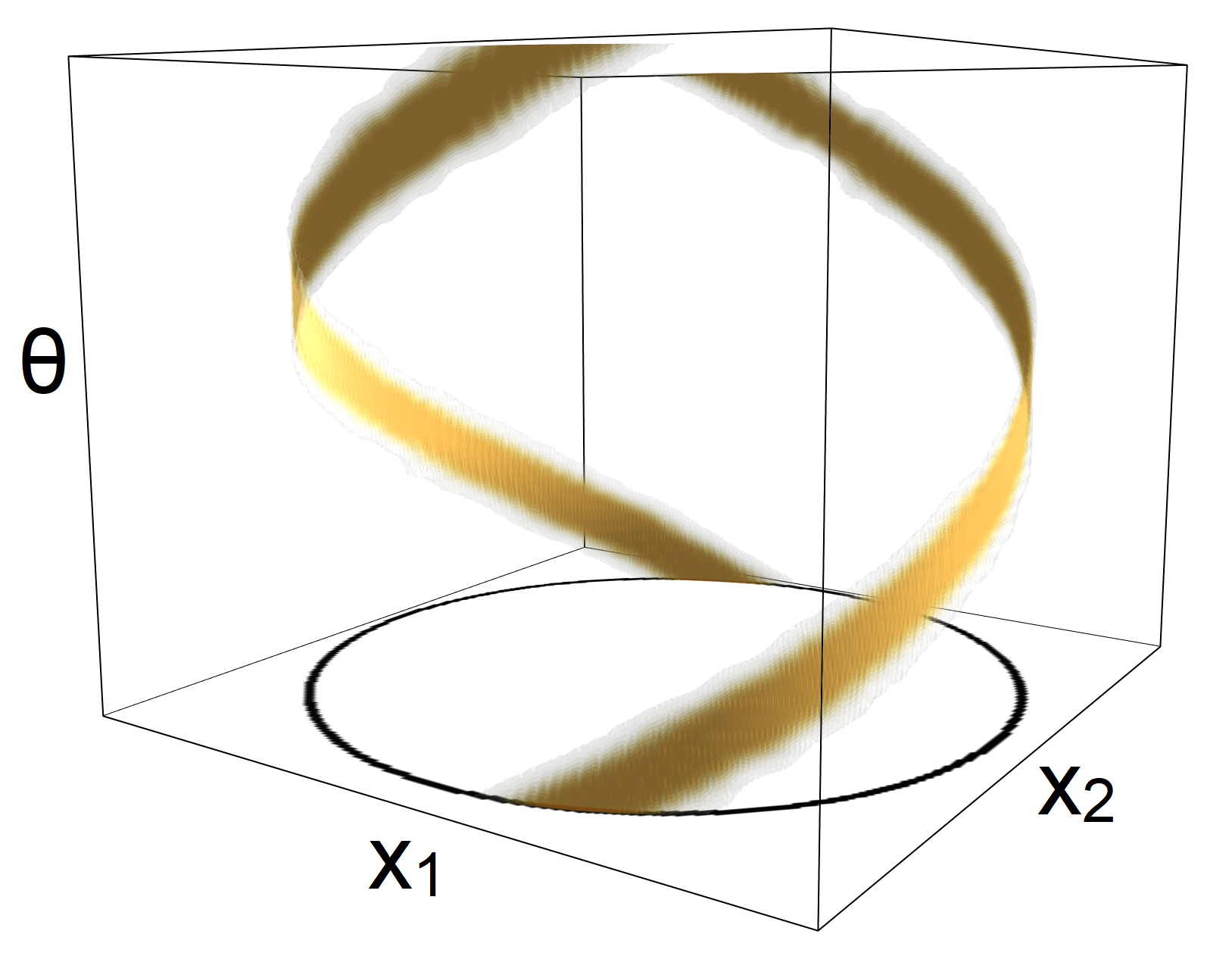}
         \caption{}
         \label{fig:os_circle}
     \end{subfigure}
    \caption{Examples of the orientation scores: a) Orientation score of an image with lines; b)  Orientation score of an image with a circle. }
    \label{fig:Examples of orientation scores}
\end{figure}

In \cite{Bekkers2014} a general image processing method was proposed for line tracking in 2D images via 3D multi-orientation distributions (so-called `orientation scores').
Such an orientation score \cite{duits2005phd}  lifts the image-domain from a 2D position space to a 3D space $\mathbb{M}_{2}$ of 2D-positions and orientations, as can be seen in ~Fig.~\ref{fig:Examples of orientation scores}.
Where a 2D image $f: \R^2 \to \R$ assigns a greyvalue $f(x_1,x_2)$ to a position $(x_1,x_2)$, an orientation score $U_f:\mathbb{M}_{2} \to \mathbb{C}$ assigns a complex value $U_f(x_1,x_2,\theta)$ to a `local orientation' $p=(x_1,x_2,\theta)$. 
The real part of $U_f$ gives the measure of alignment of the image local line structures with the specific orientation, whereas the imaginary part can be interpreted as the measure of alignment of image local edge structures with the specific orientation. 

In \cite{Bekkers2015} the tracking in the orientation scores (TOS) is done by formal optimal curve algorithms. 
The curve extraction is done with a single optimal control problem, 
allowing to include local contextual alignment models of oriented image features. 
In other words, optimality of the curve is determined not only by the intensity of pixels it lies on, but also by the alignment of this curve with the image's local anisotropic structures. 

These contextual geometric models in $\mathbb{M}_2$ improve tracking when cracks are partly visible, which is often the case.
Furthermore, cracks often have sudden changes in local orientation, and do not have cusps like in \cite{Bekkers2015}. Therefore, the used algorithm relies on the improved forward motion model also applied in \cite{van2022geodesic} for blood vessel tracking. For efficient sufficiently accurate computation anisotropic fast marching algorithm \cite{Mirebeau2012} is used to compute distance maps in $\mathbb{M}_2$. A subsequent steepest descent provides the optimal paths, i.e. optimal geodesics being paths with minimal data-driven length.  

\begin{figure}
\centering
\includegraphics[width= 0.9\textwidth]{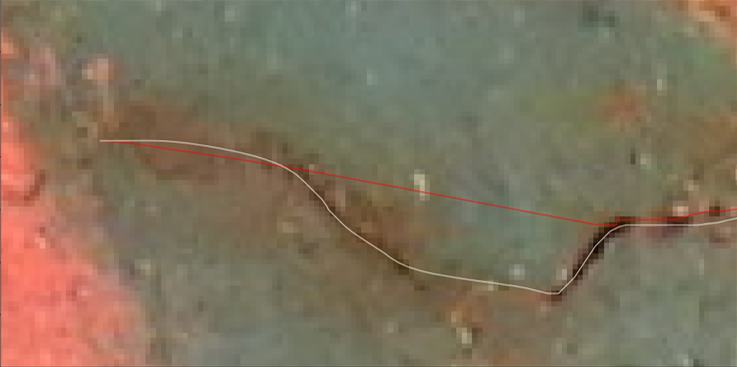}
\caption{Close up view of a crack with their calculated crack path retrieved by two methods: The red line shows the track obtained by tracking in $ \R^2$ while the white line shows the spatially projected track obtained by tracking in $\mathbb{M}_2 = \mathbb{R}^2\times S^1$.}
\label{fig:M2 R2 tracking comparison}
\end{figure}

Such geometric TOS \cite{van2022geodesic} gives a few advantages in the application of tracking cracks in steel bridges:

\begin{itemize}
\item When a crack is only partly visible
in the image as in Fig. \ref{fig:M2 R2 tracking comparison} the aforementioned contextual image processing models are needed to still recognize the crack
\item
 Crossing structures (e.g. a crack and a sub-surface due to shading/paint) are disentangled in the orientation score  (e.g. Fig.~\ref{fig:input image} and Fig.~\ref{fig:os theta levels}). 
 The crack and the edge/line crossing are separated because they align with different orientations, thus being on different ``$\theta$ levels" in the orientation score.
Thereby a tracking algorithm will stay on a true crack trail.
\item Cracks can suddenly change orientation and the model based on \cite{Duits2018_27} automatically accounts for that.
\end{itemize}

\subsubsection{Construction of the orientation score}
\label{sec:Construction of Orientation Score}

\begin{figure}
     \centering
     \begin{subfigure}[b]{0.45\textwidth}
         \centering
         \includegraphics[width=\textwidth]{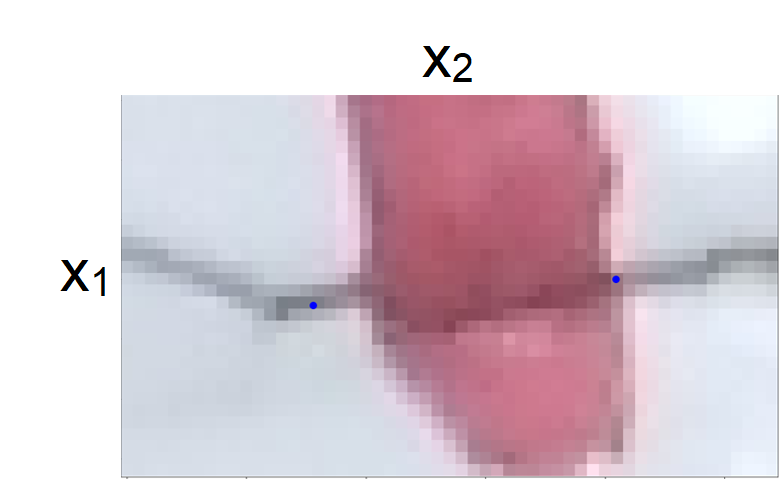}
         \caption{}
         \label{fig:input image}
     \end{subfigure}
     \hfill
     \begin{subfigure}[b]{0.45\textwidth}
         \centering
         \includegraphics[width=\textwidth]{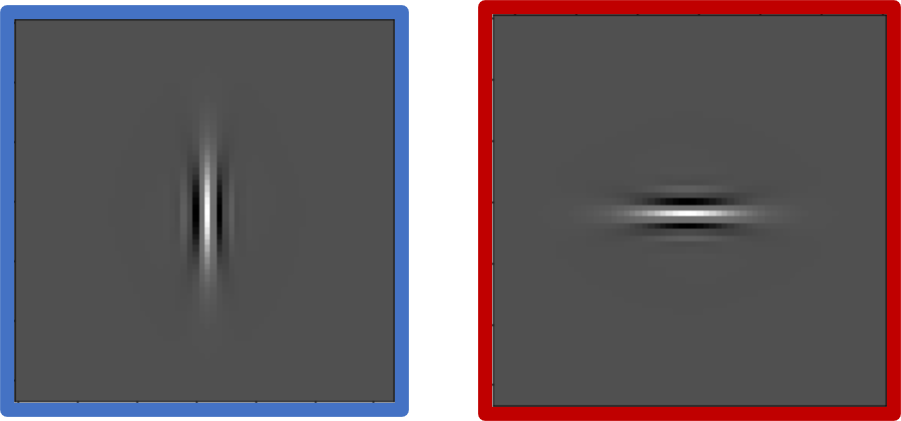}
         \caption{}
         \label{fig:cake wavelets}
     \end{subfigure}
    \caption{a) Input image with chosen endpoints; b) Real part of the cake wavelets with angle $ \theta$ equal to $0$ and $\frac{\pi}{2}$ respectively. }
    \label{fig:input image and cake wavelets}
\end{figure}
\begin{figure}
     \centering
     \begin{subfigure}[b]{0.3\textwidth}
         \centering
         \includegraphics[width=\textwidth]{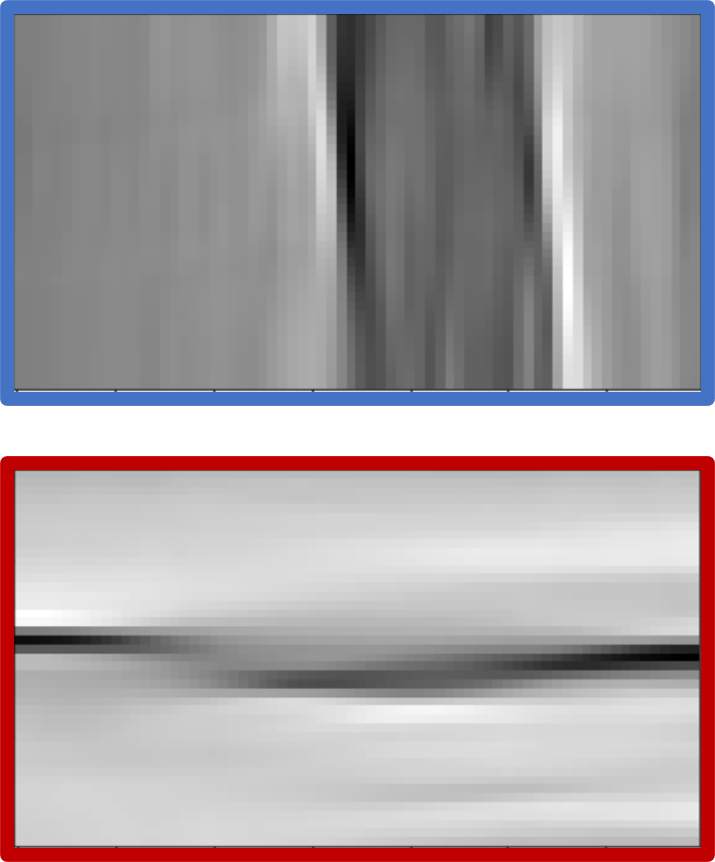}
         \caption{}
         \label{fig: theta-slice}
     \end{subfigure}
     \hfill
     \begin{subfigure}[b]{0.6\textwidth}
         \centering
         \includegraphics[width=\textwidth]{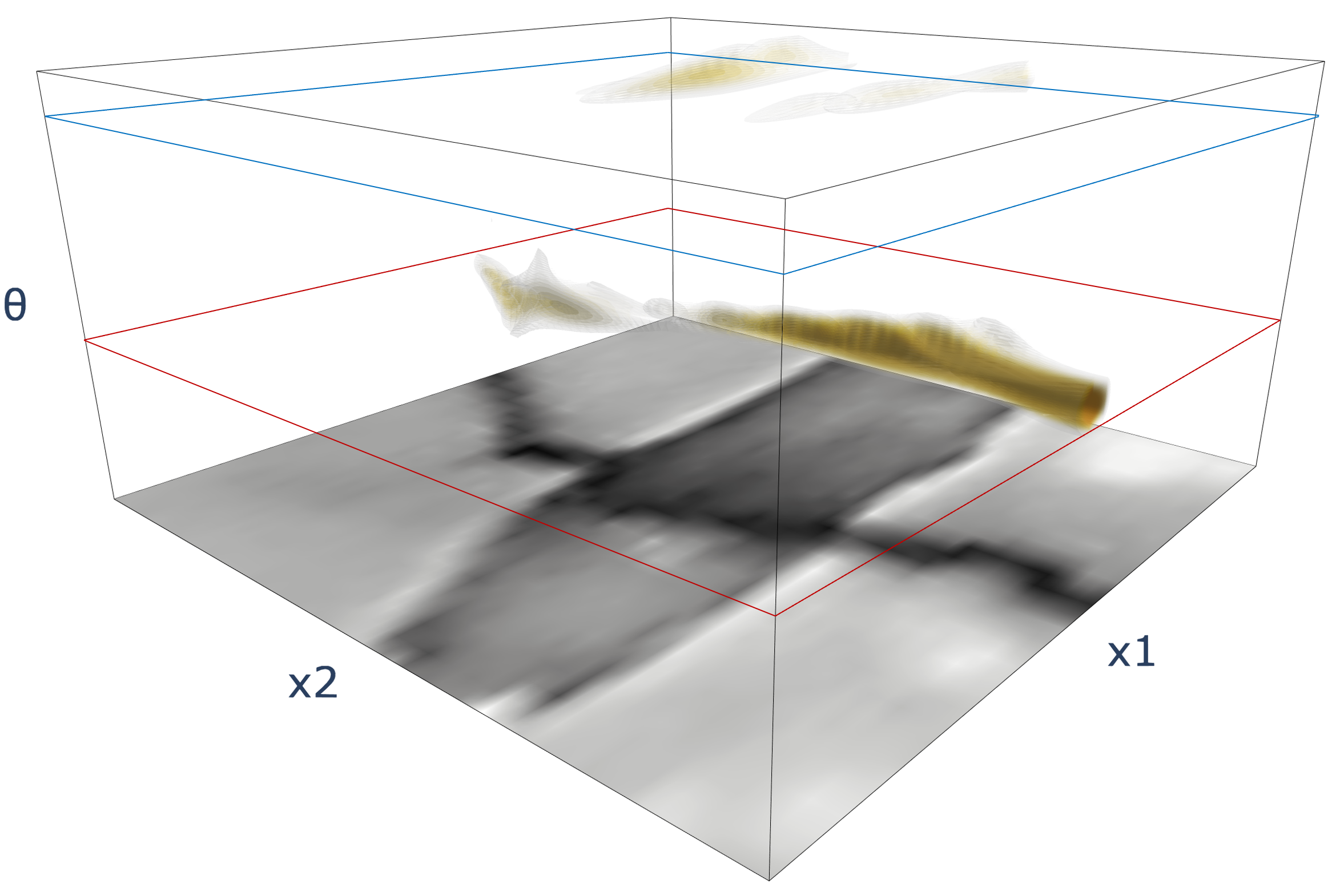}
         \caption{}
         \label{fig:os theta levels}
     \end{subfigure}
    \caption{a) Results of applying filters from figure  Fig. \ref{fig:cake wavelets} to the image shown in Fig. \ref{fig:input image}; b) Orientation score of image shown in Fig. \ref{fig:input image}. $ \theta$-levels represented by blue and red rectangles correspond to a).
    }
    \label{fig:theta-slice and os}
\end{figure}

In the orientation score, the image domain of positions $ \mathbf{x}:=(x_1,x_2)\in\mathbb{R}^2$ is extended to the domain of positions and orientations $(\mathbf{x},\theta)$, where $\theta\in[0,2\pi]$ denotes the orientation. 
The orientation score can be built from the initial image using the anisotropic cake wavelets shown in Fig. \ref{fig:cake wavelets} (which thank their name to their shape in the Fourier domain \cite{Bekkers2014,duits2005phd}). 
The convolution of an image with a cake wavelet that has a specific orientation $\theta$ will filter the local line elements, giving high response in positions where the line structures are aligned with the applied wavelet orientation. 
For example, the responses in Fig.\ref{fig: theta-slice} are obtained by applying the wavelet filters shown in Fig.\ref{fig:cake wavelets} to the image shown in Fig.\ref{fig:input image}.

A grayscale image can be considered as a (square integrable) function $f:\R^2 \to \R$ that maps a position $\mathbf{x} \in \R^2$ to a grayscale value $f(\mathbf{x})$. Similarly, the orientation score $U_f$ is represented by the function 
\mbox{$U_{f}:\mathbb{M}_2\rightarrow \mathbb{C}$}, where \mbox{$ \mathbb{M}_2 = \mathbb{ R}^{2}\times S^1$} and \mbox{$S^1:=\left\{\left.\mathbf{n}(\theta)=(\cos \theta,\sin \theta)\;\right|\; \theta \in [0,2\pi)\right\}$}. 
The used cake wavelets $\psi$ are complex-valued (their real part detects lines whereas their imaginary part detects edges). 
The orientation score $U_f$ of an image $f$ is 
given by:
\[U_f(\mathbf{x},\theta) = \int_{\mathbb{R}^2} \overline{\psi \left( R_\theta^{-1}(\mathbf{y} - \mathbf{x}) \right)} f(y)\, \mathrm{d}\mathbf{y}, \textrm{ for all } \mathbf{x} \in \R^2, \theta \in [0,2\pi),
\]
where $ \psi $ is the complex-valued wavelet aligned with an a priori axis (say the vertical axis $\theta=0$), and the rotation matrix $ R_\theta$ rotates this wavelet counterclockwise with the required angle $\theta$ and is defined by:
{\small $R_\theta = 
\begin{pmatrix}
\cos\theta & -\sin\theta\\
\sin\theta & \cos\theta
\end{pmatrix}
$}.
A careful mathematical design of the cake wavelets \cite{Bekkers2014,duits2005phd} allows to preserve all image information after ``lifting'' the image to the orientation score. 
With a proper choice of parameters that are used to construct the cake-wavelet, the approximate reconstruction of the image $f$ from the orientation score can be achieved by a simple integration: 
\[
f(\mathbf{x})= \int_{0}^{2\pi} U_f(\mathbf{x},\theta) {\rm d}\theta.
\]

\subsubsection{Shortest paths (geodesics) in the orientation score}
\label{sec:Shortest Paths (Geodesics) in the Orientation Score}
\begin{figure}
     \centering
     \begin{subfigure}[b]{1\textwidth}
         \centering
         \includegraphics[width=\textwidth]{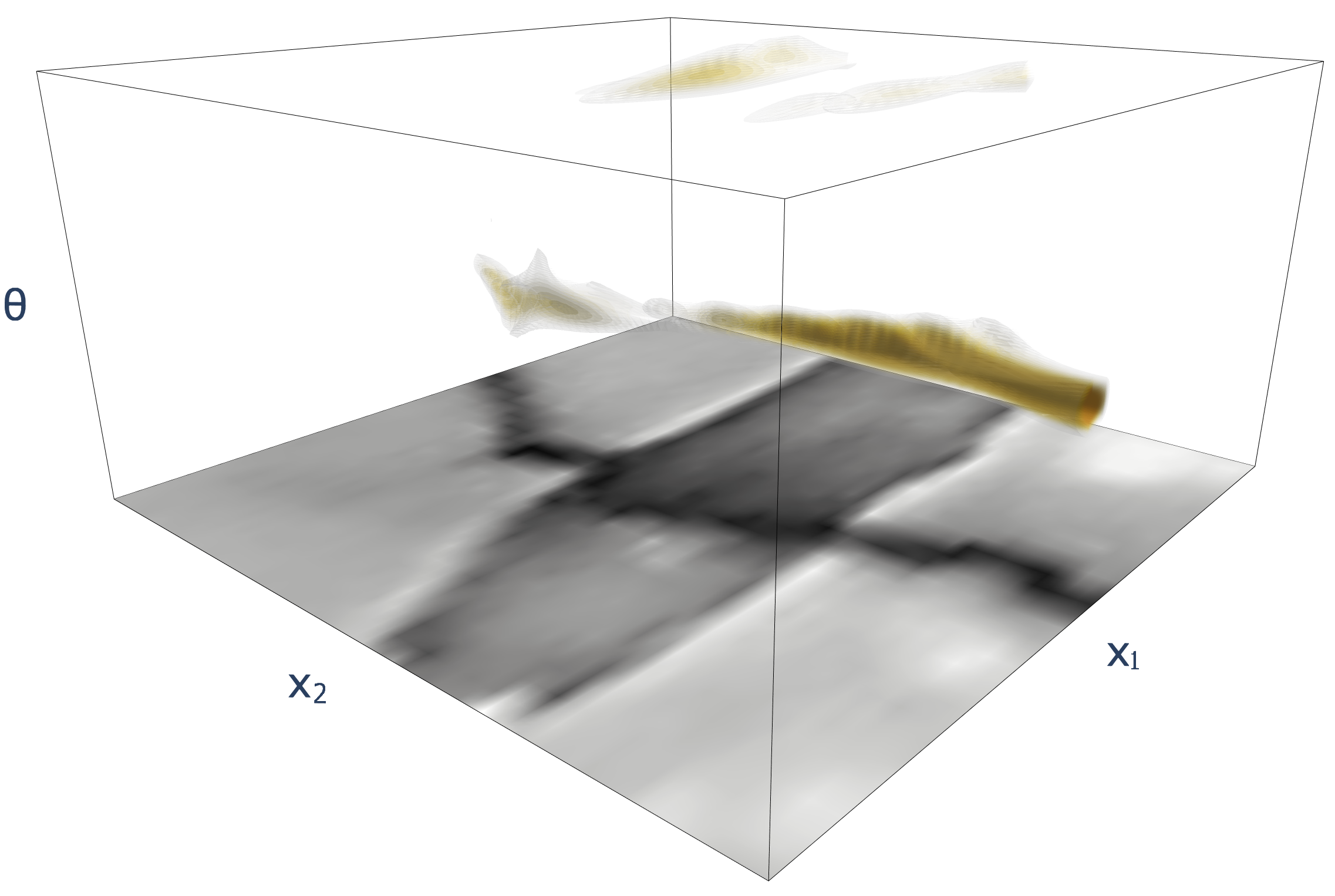}
         \caption{}
         \label{fig: os}
     \end{subfigure}
     \hfill
     \begin{subfigure}[b]{1\textwidth}
         \centering
         \includegraphics[width=\textwidth]{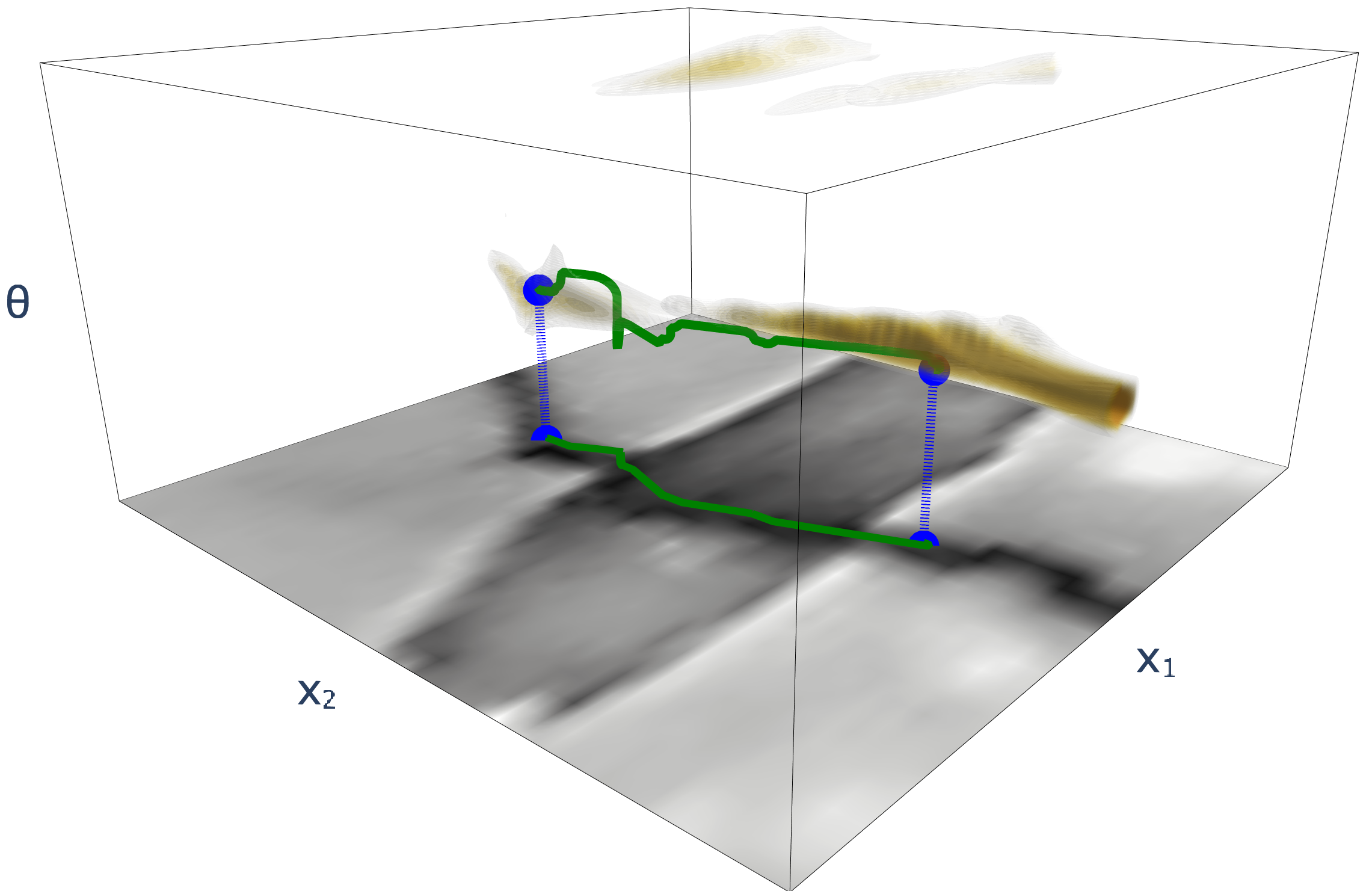}
         \caption{}
         \label{fig:track in os}
     \end{subfigure}
    \caption{a) Orientation score; b) An example of a geodesic (the shortest path $ \gamma$ from Eq. \ref{dist}) in the orientation score and its projection onto $ \R^2$.
    }
    \label{fig:Tracking in M2}
\end{figure}

After the input image is lifted to the orientation scores, a tracking algorithm is applied that finds the shortest path, or geodesic, between the endpoints of the crack in the orientation score, see Fig.~\ref{fig:Tracking in M2}. The geodesic can be represented by a parameterised curve $\gamma(t)=(\mathbf{x}(t),\mathbf{n}(t))$ in the orientation scores, the length of which is defined as the Riemannian distance between the chosen endpoints $\mathbf{p} = (\mathbf{x}_0,\mathbf{n}_0)$ and $ \mathbf{q}=(\mathbf{x}_1,\mathbf{n}_1)$. Here, $\mathbf{p}$ and $\mathbf{q}$ are points in the lifted space of positions and orientations $ \mathbb{M}_2:=\mathbb{R}^2\times S^1$ and $\mathbf{n}_t:= \begin{pmatrix}
    \cos{\theta_t} \\ 
    \sin{\theta_t} 
\end{pmatrix}$. The asymmetric version of the Riemannian distance, that were used in the experiments, is defined by:

\begin{equation} \label{dist}
d_G(\mathbf{p},\mathbf{q}) = \inf_{\substack{\gamma(\cdot)=
\left(\mathbf{x}(\cdot), \mathbf{n}(\cdot)\right) \in \Gamma\\ \gamma(0)=\mathbf{p}, \gamma(1)=\mathbf{q} \\ \dot{\mathbf{x}}(\cdot) \cdot \mathbf{n}(\cdot) \geq 0}} \int_{0}^{1} \sqrt{G_{\gamma(t)} (\dot{\gamma}(t),\dot{\gamma}(t))} \mathrm{d}t
\end{equation}
where the space of curves $\Gamma$ (are all piecewise continuously differentiable curves $\gamma:[0,1]\rightarrow \mathbb{M}_2$) over which we optimize,
and with the Riemannian metric given by {\small
\begin{equation}\label{G} 
G_\mathbf{p}(\dot{\mathbf{p}},\dot{\mathbf{p}}) \!=\! C^2(\mathbf{p})\!\left(\! \xi^2|\dot{\mathbf{x}}\cdot \mathbf{n}|^2\! + \! \frac{\xi^2}{\zeta^2} \left(\lVert \dot{\mathbf{x}}\rVert^2\!\!-\!|\dot{\mathbf{x}}\cdot \mathbf{n}|^2\right) \!+\! \lVert \dot{\mathbf{n}} \rVert^2 \!+\! 
\lambda 
{\scriptsize
\frac{\max\limits_{\|\dot{\mathbf{q}}\|=1} \left| HU|_\mathbf{p}(\dot{\mathbf{p}},\dot{\mathbf{q}})\right|^2}{\max\limits_{\substack{\|\dot{\mathbf{q}}\|=1,\\\|\dot{\mathbf{p}}\|=1}} \left| HU|_\mathbf{p}(\dot{\mathbf{p}},\dot{\mathbf{q}}) \right|^2}} \!\right)
\end{equation}}with $\mathbf{p}=(\mathbf{x},\mathbf{n}) \in \mathbb{M}_2$ being a position and orientation, and $\dot{\mathbf{p}}=(\dot{\mathbf{x}},\dot{\mathbf{n}})$ a velocity attached to the point $\mathbf{p}$, and 
where $ C^2(\mathbf{p})$ is the output of the application of the multi-scale crossing/edge preserving line filter to the orientation score, for details see \cite[App.D]{van2022geodesic}, \cite{Hannink2014}, and acts as a cost function. The Hessian of the orientation score is denoted by $ HU$. 
Parameter $\xi>0$ influences the stiffness or curvature of the geodesics. Parameter $0<\zeta \ll 1$ puts high cost on $ \xi^2/ \zeta^2$ on sideward motion relative to the cost $ \xi^2$ for forward motion. Parameter $\lambda>0$ regulates the influence of the data-driven term relying on the Hessian of the orientation scores.
In order to find the optimal curve that minimizes the distance, the anisotropic fast marching algorithm \cite{Mirebeau2012} is applied to solve the eikonal PDE \cite{Duits2018_27}. The algorithm computes the distance map $d_{G}(\mathbf{p},\cdot)$ given by Eq. (\ref{dist})
from one of the crack endpoints, $\mathbf{p}$, that serves as an initial condition (i.e. `seed point') using an efficient propagating front approach. Afterwards, it finds the geodesics by backtracking using the steepest descent method \cite{Duits2018_27} from the other endpoint $\mathbf{q}$, also called the `tip point', using the computed distance map. The HFM library \cite{HFM} was used to employ the described tracking algorithm efficiently.

Next, extra motivation for the curve optimization model given by Eq. (\ref{dist}) is provided.  When considering crack propagation as a random Brownian process \cite{DuitsAMS1,ZhangDuits} on $\mathbb{M}_2$ then Brownian bridges concentrate on geodesics \cite{wittich2005explicit}. Furthermore, the model given by Eq. (\ref{dist}) corresponds to a curvature-adaptive extension of the model in \cite{Duits2018_27} whose optimal geodesics include `in-place rotations' that the model automatically places at optimal locations during the 
geodesic distance front-propagation.
Such `in-place rotations' (due to the 
constraint $\dot{\mathbf{x}} \cdot \mathbf{n} \geq 0$
in Eq. (\ref{dist}))  are natural for 2D pictures of cracks in bridges, as there are often sudden changes in direction of a crack. 

\subsection{Crack width detection}
\label{sec:Crack width detection}
Two distinct approaches are considered that allow to obtain a crack segmentation while having the crack path.  
The first approach is called 'width expansion' (WE) and the second approach 'edge tracking' (ET).
The WE approach is expected to work better with cracks that have a conspicuous grain structure along their edges, e.g. cracks in pavement.
In contrast, the ET approach gives better results for cracks that have relatively smooth edges such as cracks in steel. 
\subsubsection{Crack width expansion}
\label{sec:Width expansion}
This approach has been used in automatic crack segmentation algorithms \cite{amhaz2014new,amhaz2016automatic}, and is applied to the image in the $ \mathbb{R}^2$ domain without any preliminary transformations. 
In the first iteration of the WE algorithm, the pixels that lie exactly on the obtained crack path are assigned to a crack segment.
In the subsequent iterations, pixels neighboring to the existing crack segment are assigned to a crack segment if their intensity value is below a threshold value $ T_w = \mu_w - K_w\cdot\sigma_w$, where $ \mu_w $ and $ \sigma_w $ are mean and standard deviation respectively of the existing crack segment pixels gray value distribution, and where $ K_w $ is a manually chosen parameter. 
The iterations continue until no new pixels are added to the crack segment.
This approach is useful when the crack has an irregular shape as is the case for e.g. cracks in pavement, due to the noticeable grains of the material.
However, the output of this approach depends significantly on the threshold value $ K_w $. 
The necessity to accurately chose $ K_w $ restricts the autonomy of the method, since, its optimal value may vary significantly depending on the image. 

\subsubsection{Crack edge tracking}
\label{sec:Crack edge tracking}
Similarly as for crack path tracking, for the ET a geometric tracking algorithm is applied to find an optimal path between crack endpoints, but with a few major differences.

\begin{figure}
     \centering
     \begin{subfigure}[b]{0.45\textwidth}
         \centering
         \includegraphics[width=\textwidth]{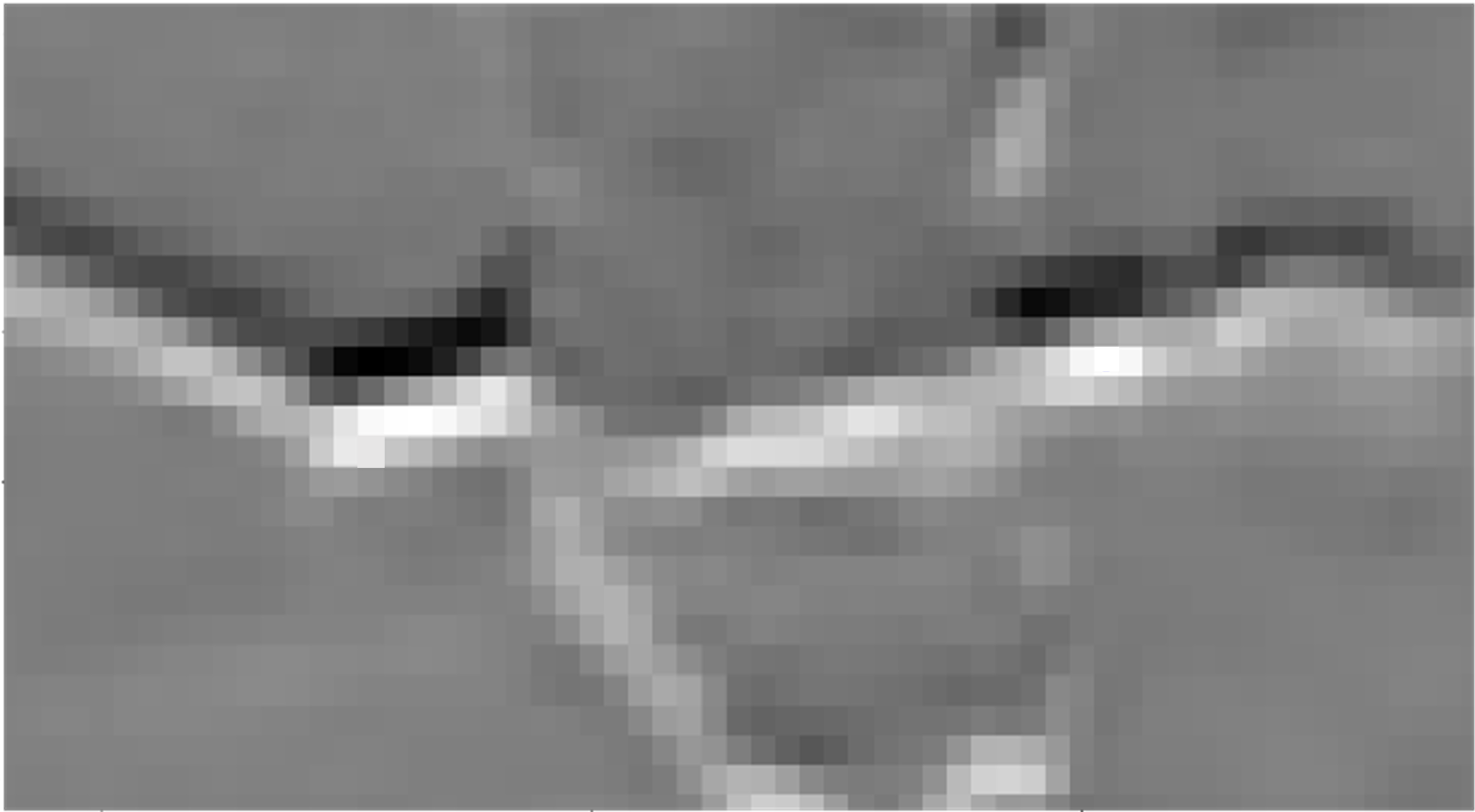}
         \caption{}
         \label{fig:Edge cost function}
     \end{subfigure}
     \hfill
     \begin{subfigure}[b]{0.45\textwidth}
         \centering
         \includegraphics[width=\textwidth]{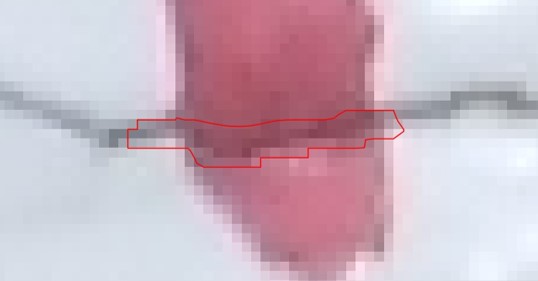}
         \caption{}
         \label{fig:Edge tracks}
     \end{subfigure}
    \caption{a)  Oriented edge filter applied to the image shown in the Fig. \ref{fig:input image}; b) Crack edges detected with the ET method.
    }\label{fig:Edge tracking}
\end{figure}

First of all, a cost function $C(\mathbf{p})$ for the tracking algorithm is built that forces it to follow the crack edges instead of the crack centerline. 
To build this cost function, the crack path is divided into small segments for which crack path orientation is determined. Afterwards, around each path segment, a square part of image is chosen with a predetermined dimensions in parallel and perpendicular to the segment orientation. 
To these pieces of the image, an edge filter (Gaussian first order derivative) oriented in accordance with the crack path orientation is applied. Results of this operations are summed to construct a cost function for the whole image. 
In this way, the edge filter highlights mainly the edges of the crack and gives lower responses to edges not parallel to the crack.
An example of a resulting image is presented in Fig. \ref{fig:Edge cost function}.


Additionally, in the proposed ET method we use a version of the tracking algorithm described in Section \ref{sec:Shortest Paths (Geodesics) in the Orientation Score} that in fact is the Dijkstra's algorithm in the image $ \R^2$ domain. 
As a filter was applied that was specifically designed to highlight crack edges and ignore other edges on the image, it is expected to avoid the problems listed in Section \ref{sec:Crack path tracking} (namely crossing structures and poor visibility of the filtered crack edges).
Hence, instead of doing tracking in the $\mathbb{M}_2$ domain as was done for crack path tracking, here the fast marching algorithm  \cite{Mirebeau2012} for geodesic curve optimisation is directly applied in the image  $\R^2$ domain that is computationally less expensive.



\section{Results}
\label{sec:Results}

\subsection{Tracking performance}
\label{sec:Tracking performance}
First, the performance of the crack path tracking algorithm is evaluated against the steel structures dataset. The used dataset consists of images of actual steel bridge structures with fatigue cracks that was collected by Dutch infrastructure authorities. Ground truth crack tracks (actual track of a crack as it is visible on the image) were drawn manually for 19 randomly selected images from the dataset. 
In order to numerically estimate the performance of the tracking method, the metric $ m =  \frac{A}{L^2}$ is introduced, 
where $A$ is the area between the ground truth track and retrieved track, measured in squared pixel, and $L$ is the length of the ground truth track determined as the number of pixels it passes. 
The area $A$ between to tracks represents a measure of how one track deviates from another.
Division by $ L^2$ makes the metric resolution-invariant. 
Lower values of the metric $ m$ mark lesser deviation of the retrieved crack from the ground truth track.
Table \ref{tab:tracking results} shows the mean value of the introduced metric.

The described tracking method was compared with the alternative methods.
Table \ref{tab:tracking results} shows the results of this comparison, where the considered algorithms are identified as:
\begin{itemize}
\item FF --- the "FlyFisher" algorithm described in \cite{dare2002operational}. In this method a track propagates a track from the starting point and led by the local gray-scale pixel values patterns;
\item DT --- Tracking 2D paths using Dijkstra's algorithm in the $  \R^2$ domain of the input image to minimize the cost function defined by the Frangi filter~\cite{frangi1998multiscale};
\item TOS --- Tracking in the Orientation Score: Tracking optimal paths defined by Eq. (\ref{dist}) in the orientation score via anisotropic fast marching algorithm (described in Subsection \ref{sec:Crack path tracking}).
\end{itemize}
\newcommand\ChangeRT[1]{\noalign{\hrule height #1}}
\begin{table}
\centering
\caption{ Average deviation of three different crack tracking algorithms from the ground truth crack on 19 images of cracks in steel structures.}
\label{tab:tracking results}

\begin{tabular}{m{0.2\textwidth} m{0.2\textwidth} m{0.2\textwidth} m{0.2\textwidth}}
\setlength{\arrayrulewidth}{10mm}
\\
\hline
  & FF & DT & TOS \\
\hline
$ m\cdot10^5$ & 4.46 & 0.980948 &	0.784308 \\
\hline
\end{tabular}
\end{table}
The results demonstrate that the proposed TOS method outperforms DT and FF algorithms. 
The path retrieved by the FF method follows a locally optimal direction (defined by a darker image region) but easily misses a globally optimal path.
This problem results in the worst performance (see Table \ref{tab:tracking results}). 
In contrast, the TOS and the DT algorithm use a two step approach. 
First, they calculate the distance map on a Riemannian manifold $(M, G)$ 
from the 
starting point  
and then perform a backtracking using the steepest descent algorithm. In case of TOS, the manifold $M=\mathbb{M}_{2}$
and metric tensor field $G$ given by Eq.~(\ref{G})
while in the DT algorithm, the base manifold $M=\R^2$ and the metric tensor field $\mathcal{G}$ 
is now given by $\mathcal{G}_{\mathbf{x}}(
\dot{\mathbf{x}},\dot{\mathbf{x}})= V(\mathbf{x}) \|\dot{\mathbf{x}}\|^2$ with $V$ the Frangi filter \cite{frangi1998multiscale}.

This two step approach allows us to find the optimal path resulting in significantly lower values of the metric $ m$ in Table \ref{tab:tracking results}.
Here the DT algorithm follows the path with the lowest cumulative Frangi filter values $V$-intensity of pixels.
The TOS algorithm improves upon this by using a crossing-preserving $V$- intensity \cite{Hannink2014}, and moreover it includes alignment of locally  oriented image features as explained in Section \ref{sec:Crack path tracking}, resulting in better tracking performance (see Table~\ref{tab:tracking results}).
\subsection{Segmentation on the AigleRN dataset}
\label{sec:AigleRN dataset}
The AigleRN dataset contains images of cracks in pavement and provides ground truth crack segmentations \cite{AigleRN}.
In order to segment cracks on images from this dataset, the TOS method is used with the WE as described in Section \ref{sec:Width expansion}.
Seventeen images were randomly selected from the AigleRN dataset for the evaluation. 
To evaluate the performance of the tracking algorithm, criterions were used as in \cite{chen2021improved}, namely precision, recall and F1-value.
Recall shows what fraction of the crack pixels were retrieved by an algorithm. 
Recall equal to 1 means that all pixels that belong to a crack according to ground truth were identified by an algorithm as crack pixels. 
The precision value indicates what fraction of the pixels that were identified as crack by an algorithm, actually belong to a crack segment. Finally, the F1-value is the harmonic mean of precision and recall.
Results are provided in Table \ref{tab:AigleRN results}. 

In \cite{AigleRN} crack segmentation results retrieved by minimal path selection (MPS) \cite{amhaz2016automatic} and free-form anisotropy (FFA) \cite{nguyen2011free} algorithms are also provided. 
Using these provided segmentation results we also evaluated the performance of the MPS and the FFA methods which are also shown in Table \ref{tab:AigleRN results} for the same seventeen images. 
\begin{table}
\centering
\caption{Crack segmentation results on the AigleRN dataset.}\label{tab:AigleRN results}
\begin{tabular}{m{0.2\textwidth} m{0.2\textwidth} m{0.2\textwidth} m{0.2\textwidth}}
\setlength{\arrayrulewidth}{10mm}
\\
\hline
 &   FFA & MPS & TOS+WE\\
\hline
Precision & 0.61 & 0.82 &	0.91 \\
Recall & 0.32 & 0.6 & 0.75\\
F1 & 0.36 & 0.63 & 0.82\\
\hline
\end{tabular}
\end{table}

The FFA algorithm measures a local texture anisotropy for every pixel of the image. 
The pixels where an anisotropy is higher than a threshold value are assigned to a crack segment.
This approach has the lowest performance.

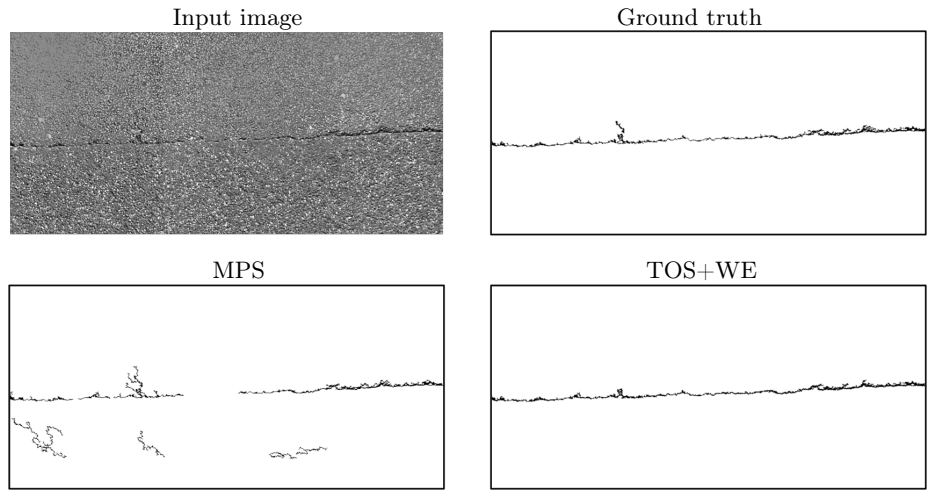
\begin{figure}

\centering
\includegraphics[width = 1\textwidth]{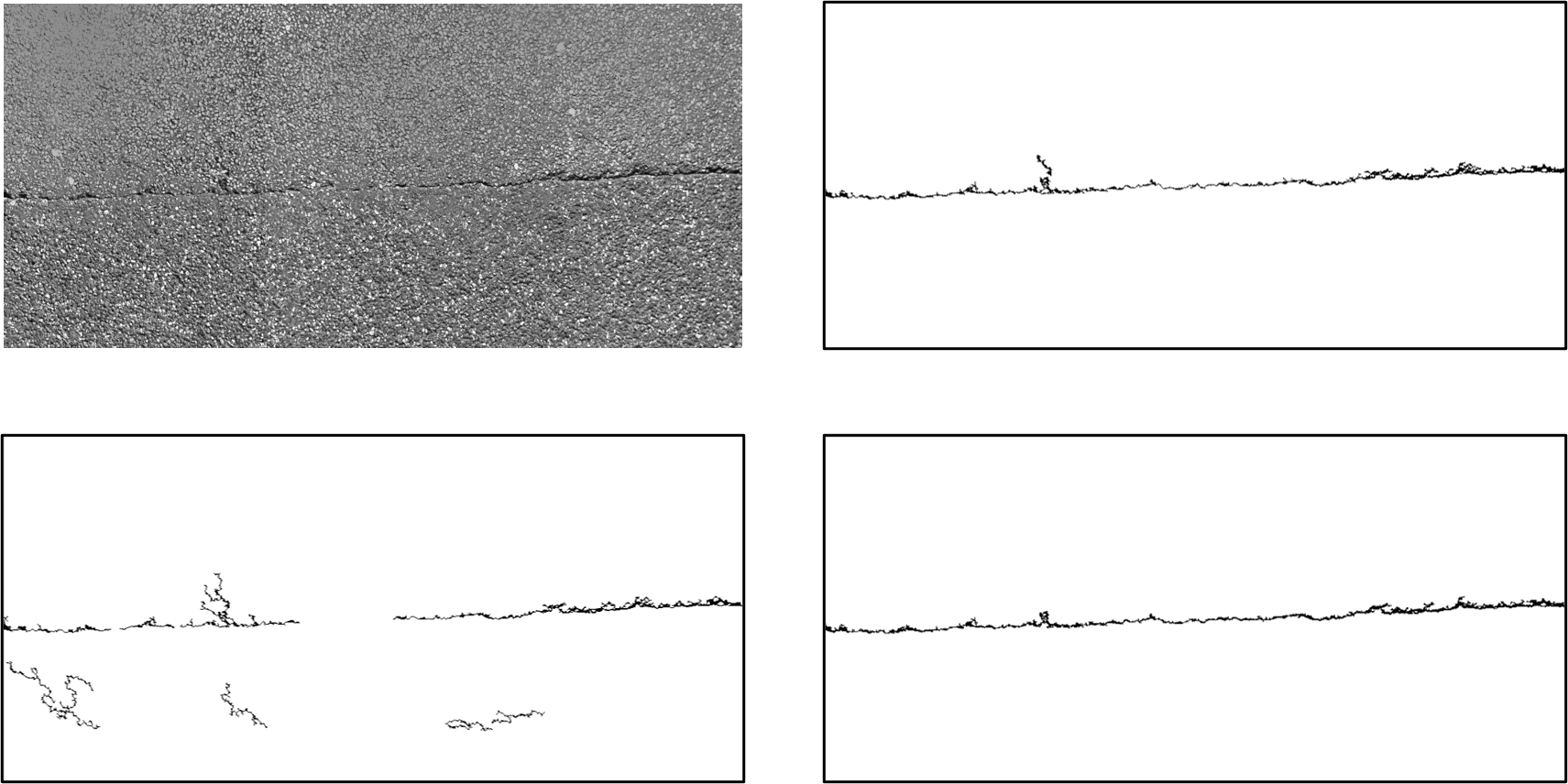}
\put(-285,176){Input image}
\put(-117,176){Ground truth}
\put(-270,80){MPS}
\put(-106,80){TOS+WE}
\caption{Typical example in the AiglRN dataset with segmentations. 
The output of the TOS+WE algorithm is more connected and closer to ground truth than the MPS-output. This qualitatively supports the performance differences in~Table~\ref{tab:AigleRN results}. }
\label{fig:AigleRN}
\end{figure}

The MPS algorithm finds potential crack pixels using a threshold of local  minima in grayscale. 
Afterwards, the paths between these points are computed using Dijkstra's algorithm. 
These paths constitutes the so-called `skeleton' of the crack. 
To find crack segments, this skeleton is used as a basis for the WE algorithm as explained in Section \ref{sec:Crack width detection}.
The MPS algorithm does not allow to detect cracks with poor visibility, and this explains the low recall value. 
This can also be observed in Fig. \ref{fig:AigleRN}, where the middle part of the crack with partial visibility was not detected by the MPS method. 
Also, in Fig. \ref{fig:AigleRN} we can observe false detections by the MPS algorithm, whereas the TOS+WE algorithm avoids false detections and maintains the connectivity of the crack. Because of this effects the proposed TOS+WE algorithm outperforms its two counterparts by all three metrics as can be seen on Table \ref{tab:AigleRN results}.


\subsection{Segmentation on steel structures images}
\label{sec:Steel structures dataset}
Finally, results are provided of the application of the segmentation algorithm to the images of steel bridge structures with cracks from the dataset introduces in Section \ref{sec:Tracking performance}.
Ten images with visible crack edges from the dataset were used for the evaluation. 
Unlike in pavement, crack edges in steel structures have a smooth shape, and previously it was stated that the ET algorithm should work better than the WE algorithm in this case. Here this is shown by comparing the TOS+ET and TOS+WE algorithms in Table \ref{tab:Steel structures dataset}. 

\begin{figure}
\centering
\includegraphics[width = 1\textwidth]{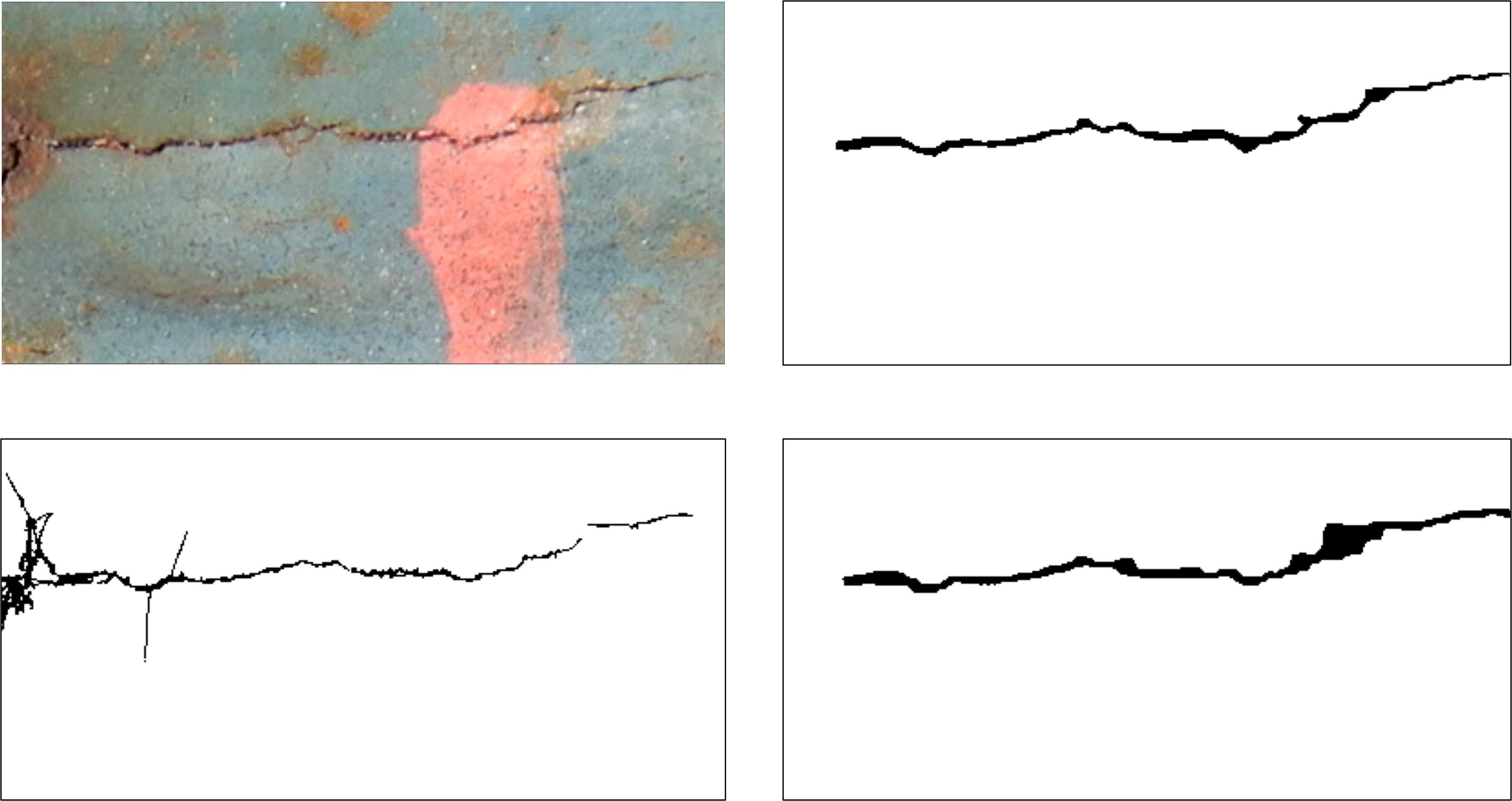}
\put(-286,186){Input image}
\put(-115,186){Ground truth}
\put(-270,86){MPS}
\put(-105,86){TOS+ET}
\caption{ Example of a steel bridge image and the corresponding results of its segmentation. The TOS+ET method suffers less from neighboring/crossing structures (paint, rust) because of the multi-orientation decomposition.}
\label{fig:steel structures}
\end{figure}
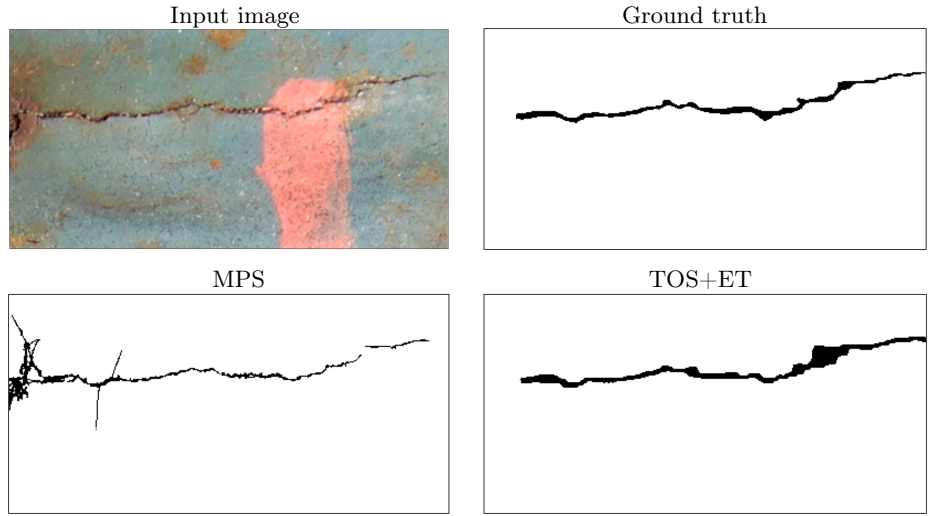

In Table \ref{tab:Steel structures dataset}, also the results for the MPS algorithm are added for comparison.
As was explained in the introduction of this paper, the fully automatic MPS method tends to have lots of false crack pixels identifications on images with non-crack dark elongated elements as can be seen on Fig. \ref{fig:steel structures}.
This leads to only 49\% precision for the MPS algorithm and low F1-value.

In contrast, with the proposed algorithms (TOS+WE and TOS+ET) the image structure of interest is identified at first using the crack path tracking. 
Thus, the precision of the proposed algorithms is higher than that of the MPS algorithm.

Furthermore, TOS+WE method has a slightly higher precision value than the TOS+ET method, but significantly lower recall value meaning that TOS+WE algorithm skips more crack pixel, marking them as background pixels. Thereby, by the F1-value the TOS+ET method shows the best performance on the images of steel structures from our dataset.

\begin{table}

\centering
\caption{Crack segmentation results on the dataset of images of steel structures.}
\label{tab:Steel structures dataset}
\begin{tabular}{m{0.2\textwidth} m{0.2\textwidth} m{0.2\textwidth} m{0.2\textwidth}}
\\
\hline
  & MPS & TOS+WE & TOS+ET\\
\hline
Precision & 0.49 & 0.9 & 0.86 \\
Recall & 0.66 &0.69 &  0.85\\
F1 & 0.52 & 0.78 & 0.83\\
\hline
\end{tabular}
\end{table}

\section{Conclusion}
An algorithm for the crack segmentation was proposed, consisting of two major parts. The first part of the algorithm measures a crack track between manually selected crack endpoints. The second part of the algorithm performs the crack segmentation using a crack track obtained in the first part as a basis. The crack tracking part of the algorithm adapts the approach initially developed for the analysis of vascular systems on retinal images where an efficient fast marching algorithm performs line tracking on the image lifted into space of positions and orientations. For the width measurement part of the algorithm the novel approach was proposed for the crack edge tracking and it was compared with the width expansion approach that was used in earlier works.

The proposed semi-automatic algorithm allows for a crack segmentation of images with improved accuracy compared to the fully automatic algorithms, such as minimal path selection and free form anisotropy algorithms and resolves the problem of high false detections which is inherent to fully automatic crack detection image processing algorithms.
While sacrificing autonomy of the segmentation algorithm, the accuracy significantly increases.


Potentially the developed algorithm may be used as a data labeling tool, to label images with cracks to train a machine learning algorithm. 
However, results show that the proposed algorithm does not fully reproduce the manual pixel-wise labeling.
It should be studied how much this deviation affects the performance of the machine learning algorithm trained on the data labeled with the semi-automatic segmentation tool.
Also, the algorithm can find its use in cases when it is necessary to measure a crack's geometry (length, width, curvature etc.) when the location of this crack is known, e.g. for research or for the purpose of more accurate inspection.


\section*{Acknowledgement}
Authors would like to thank the Dutch bridge infrastructure owners "ProRail" and "Rijkswaterstaat" for their support. The research is primarily funded by the Eindhoven Artificial Intelligence Systems Institute, and partly by the Dutch Foundation of Science NWO (Geometric learning for Image Analysis, VI.C 202-031).

{\small
\bibliographystyle{unsrt}
\bibliography{references}
}

\end{document}